# 11

# Diversity and Inclusion in Artificial Intelligence

By Didar Zowghi, Francesca da Rimini

To date, there has been little concrete practical advice about how to ensure that diversity and inclusion considerations should be embedded within both specific Artificial Intelligence (AI) systems and the larger global AI ecosystem. In this chapter, we present a clear definition of diversity and inclusion in AI, one which positions this concept within an evolving and holistic ecosystem. We use this definition and conceptual framing to present a set of practical guidelines primarily aimed at AI technologists, data scientists and project leaders.

Our focus is socio-technological rather than relying on purely technical or human factors. In this chapter, we use "socio-technical" to cover "how humans interact with technology within the broader societal context" [1]. A socio-technical perspective on diversity and inclusion in AI and the underlying issue of bias requires processes and procedures that involve stakeholders and end users, examine cultural dynamics and norms, and evaluate, monitor, and respond to societal impacts.

We do not claim completeness or that the identified challenges and proposed guidelines are exhaustive. Instead, we have distilled relevant information from key reports, literature reviews, grey literature findings, and informal communication with specialists to suggest representative or indicative guidelines.

CSIRO Data61's multidisciplinary and diverse team has researched the topic of diversity and inclusion in Artificial Intelligence (DI-AI) in 2022. Our iterative approach examined the findings from both our own Systematic Literature Review on DI-AI and others on related topics including Fairness, Trust, Risk and Ethics, and an open-ended grey literature search on bias, diversity, and inclusion in AI. We noted recent work by small prominent organizations, groups, networks, and thought leaders in the DI-AI space. We have presented our draft definition and framework to both Australia's National Artificial Intelligence Centre's Think Tank on Diversity and Inclusion in AI and an industry organization and incorporated their feedback.

A lodestone was the World Economic Forum's 2022 report *A Blueprint for Equity and Inclusion in Artificial Intelligence* [2]. It takes a comprehensive holistic and practical approach to how equity and inclusion should be considered both at the governance and development levels throughout and beyond the development lifecycle, applying core principles and methods from the Inclusive Design and Human-Centered Design fields to the AI ecosystem. Also important was the National Institute of Standards and Technology's report *Towards a Standard for Identifying and Managing Bias in Artificial Intelligence* [1]. It integrates findings from an extensive literature review, opinions from experts in AI bias, AI fairness and socio-technical systems, workshop outcomes and public commentary on the draft.

The socio-technical theory posits that the design and performance of all organizational systems are understood and improved only if both social and technical facets are considered interdependent parts of the complex system in the context of use. The socio-technical system refers to those systems that exhibit a complex interaction between humans, systems, and the environment where the system is situated and used. AI systems as socio-technical systems are built through a complex process beyond their mathematical and algorithmic constructs. It is well understood that computational models and algorithms cannot adequately represent and describe all the societal impacts of AI. It is the values and behaviors of humans, teams, organizations, and society that inform the design, development, and deployment of AI systems in the context of their use.

When the field of Artificial Intelligence was first established at the *Dartmouth Workshop*, even though it was largely hosted by mathematicians, the participants included multiple psychologists, cognitive scientists, economists, and political scientists. This signifies the multi-disciplinary nature of AI as a field of research at the outset. Likewise, in modern-day AI systems, the need for the commitment of all stakeholders and the active participation of society is well understood. It is imperative for AI researchers and practitioners to acquire sufficient knowledge of the societal and individual implications of AI systems and understand how different humans use and live with AI systems across cultures.

The overall aims of this chapter for the readers are:

- To understand the holistic, socio-technical, and evolving nature of Artificial Intelligence.

- To have a clear and concise definition of diversity and inclusion in AI that can be adapted and used to suit the different projects, stakeholders and use contexts.
- To have adaptable and customizable DI-AI guidelines that indicate what, how, when, by whom and where diversity and inclusion issues should be considered.

This chapter is organized into several sections. In the next section, we describe the importance of DI-AI and the potential consequences if it is neglected. We then provide the definition of DI-AI followed by the specific guidelines organized into sections in accordance with the pillars of the definition. We complete the chapter with a conclusion.

# Importance of Diversity and Inclusion in AI

AI system stakeholders in recent years have begun paying more attention to diversity and inclusion in AI concerns. There are a few reasons for this. First, the expanding body of high-level frameworks and principles emanating from government, inter- and intra-governmental agencies, businesses, not-for-profit institutions, and academia [3]. These governance resources are often produced by collaborative and consultative teams, networks, and processes, encompassing a diversity of knowledge, disciplines, and perspectives. Second, some countries and states have passed, or are in the process of drafting, legislation to monitor and control the development and deployment of specific AI applications such as predictive policing and facial recognition technology. Such legislation is driven by privacy concerns, human rights issues, and diversity, inclusion, and equity concerns [4-7]. Third, there is growing public awareness arising from mainstream media and social media coverage of the prevalence of AI systems throughout society. Problems and failures of such systems that lead to unfair, unjust, or adverse outcomes have been exposed, helping to create general perceptions that people impacted directly or indirectly by AI systems may experience a lack of agency.

However, the literature reveals that AI projects do not consistently or adequately address concerns about bias, equity, diversity, and inclusion. The reasons are varied. First, the lack of practical and customizable tools operationalizing high-level principles and guidelines such as checklists, definitions, design pattern templates, questionnaires, and requirements guidelines. Second, confusion or ambiguity about who is responsible for diversity and inclusion in the AI development process. This includes both overall responsibility and oversight of diversity and inclusion considerations in an AI project and responsibility for specific measurable objectives in discrete project stages.

If diversity and inclusion in AI are neglected, this can cause negative impacts on the AI ecosystem and slow down the adoption of AI. The most serious impacts include material harm to users of those systems, whether this is unjustified bad credit ratings, diminished education and employment opportunities, inaccurate medical diagnoses, or unwarranted criminal arrests. By understanding why diversity and inclusion in AI are critically important, project teams and stakeholders are better equipped to identify, monitor, and mitigate risks, barriers, obstacles, and challenges. Likewise, more informed and AI-literate citizens can better express their agency in individual and collective decision-making about their use of and participation in AI systems whether these be in domestic contexts (e.g., voice recognition systems), industry/corporate (e.g., social media recommendation systems), or government (e.g., laws constraining the use of open-street recognition technologies).

## Definition of Diversity and Inclusion in Artificial Intelligence

**Diversity and Inclusion in Artificial Intelligence (AI)** refers to the 'inclusion' of humans with 'diverse' attributes and perspectives in the data, process, system, and governance of the AI ecosystem.

*Diversity* refers to the representation of the differences in attributes of humans in a group or society.

*Attributes* are known facets of diversity including the protected attributes in Article 26 of the International Covenant on Civil and Political Rights (ICCPR): *race, colour, sex, language, religion, political or other opinion, national or social origin, property, birth or other status*; and (given the non-exhaustive nature of Article 26, attributes explicitly protected under Australian discrimination federal law, including but not limited to:) *age, disability, criminal record, ethnic origin, gender identity, immigrant status, intersex status, neurodiversity, sexual orientation*; and intersections of these attributes.

*Inclusion* is the process of proactively involving and representing the most relevant humans with diverse attributes; those who are impacted by, and have an impact on, the AI ecosystem context.

***AI ecosystem*** refers to the collection of 5 pillars (humans, data, process, system, and governance), plus the environment (application or business domain) within which the AI system will be deployed and used.

This definition captures the essential 5 pillars (components) of AI systems, their development and deployment in a particular environment (application or business domain) within the AI ecosystem. We advocate that Diversity and Inclusion principles should be at the center of the AI ecosystem and must be embedded in all pillars for the specific context of AI use. Below we provide a short description of each pillar.

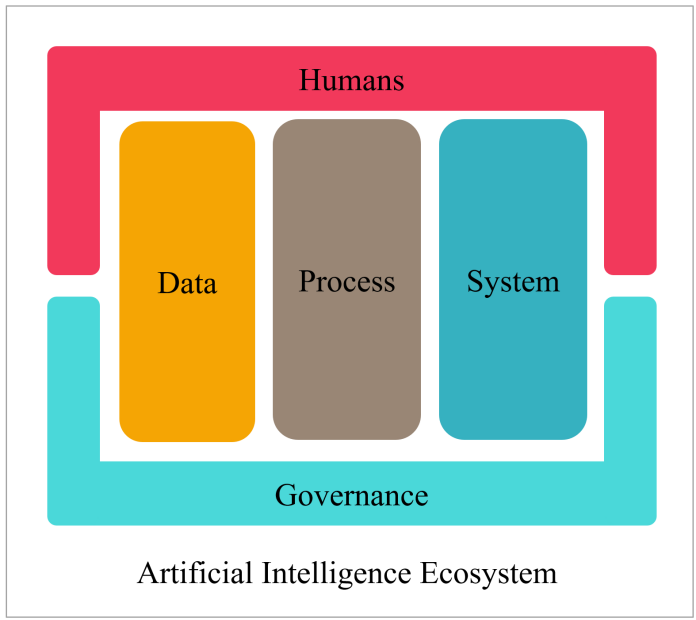

**Figure 11.1** *Pillars of Diversity & Inclusion in AI Ecosystem.*

### *Humans*

Humans are considered the core pillar of the AI ecosystem. Human-centeredness is achieved not only through the meaningful inclusion of relevant humans with diverse attributes in the building, using, monitoring, and evolution of AI systems, but also through their active participation and contribution in all the decision-making points of the AI system lifecycle. Two broad groups have been identified in the AI system lifecycle: those who will receive and use the AI system and those who will design, develop, and deploy AI systems to satisfy specific stakeholder needs. Those humans whose knowledge, lived experiences and insights are essential, need to be carefully identified, contacted, and engaged within all the relevant parts of the process. Integrating diversity and inclusion principles and practices throughout the AI system lifecycle has an important role to play in achieving equity for all humans.

### *Data*

Data plays an essential role in AI systems since it is typically through very large historical datasets that AI algorithms learn and find patterns to deliver predictions and automate decisions. What, how, why, by whom and for whom data is collected, labelled, modelled, stored, and applied has many diversity and inclusion implications. Positive and negative biases are present in the large datasets and algorithmic processes used in the development of AI models. Unwanted data biases often arise when algorithms are trained on one type of data and cannot extrapolate accurately beyond those data. In other types of AI systems (known as Symbolic AI), small datasets are used both as input and validation points in building and improving knowledge-based systems. It is critical to have a fair and inclusive representation of everyone who will be impacted by AI without any unwanted or negative bias that leads to discrimination and harm.

### *Process*

The development process describes all the activities and tasks that are carried out to deliver an AI system for a specific context of use. This definition divides the process into three sub-processes: pre-development, during development, and post-development. The **Pre-development** phase refers to the ideation of a use case or a problem that the AI system is intended to address. It also includes clearly defining the use case or the problem and the rationale behind the application of AI solution as well as identifying relevant stakeholders and eliciting their requirements. During **Development** refers to the team partnering with stakeholders to work on data collection and preparation, model design and development, and testing and evaluation of the AI system iteratively and incrementally. **Post-development** refers to the deployment of the AI system in the context of its use, monitoring its performance, safety, reliability, and trustworthiness during use, as well as making changes as necessary during the AI system lifecycle. Diversity and inclusion principles should be carefully considered and embedded throughout the entire AI system development process.

*System*

An AI system is a computer-based system that for a given set of human-defined objectives, typically uses large historical datasets to make predictions, recommendations or decisions for human consumption that may have an influence in real or virtual environments. There are many techniques and methods for verifying, validating, and monitoring AI systems (e.g. testing, algorithmic analysis of models etc), against diversity and inclusion in AI principles. AI systems must be evaluated, tested, and monitored in the context of their use to ensure non-inclusive behaviors are identified and fixed during AI system evolution. Non-adherence to practices of diversity and inclusion in the building, deployment, and use of AI systems have been shown to cause digital redlining, discrimination, and algorithmic oppression, leading to AI systems being perceived as untrustworthy and unfair.

*Governance*

AI Governance is defined as a collection of structures, processes, and regulatory and risk management frameworks that are utilized to ensure the development and deployment of AI systems are compliant with laws and regulations and conform with standards, policies, and AI Ethics principles. This definition focuses on AI governance specifically for conformance with diversity and inclusion principles. The governance component can be structured at the team, organization, and industry levels. Legal and risk frameworks should be developed and applied to guide inclusive practices in the AI ecosystems. Governance structures must be human-centered to ensure the delivery of inclusive, reliable, safe, secure and trustworthy AI systems.

# Guidelines for Diversity and Inclusion in Artificial Intelligence

Few guidelines exist to help AI practitioners ensure that diversity and inclusion considerations are embedded throughout the AI Lifecycle (AI-LC) and after deployment. Both the AI ecosystem and the AI-LC are continually evolving. Technology, governance structures, legislation, impacts, knowledge, and community expectations are never static. Accordingly, we present a series of guidelines by taking a holistic and systematic approach and mapping them to the 5 pillars of diversity and inclusion in the artificial intelligence ecosystem as in our definition above.

In some cases, regarding diversity and inclusion, literature also refers to 'equity' but for the purposes of this chapter, we exclude this term. For simplicity, we focus strictly on diversity and inclusion in AI and exclude related well-documented topics such as Responsible AI, AI Fairness, Explainable AI, AI Transparency, and AI Ethics. Moreover, although research on Bias in AI has informed our guidelines to some extent, we deemed it to be out of scope because this field has produced a plethora of detailed guidelines to counter different types of bias in different AI systems. For better referencing, we number the concrete guidelines with the first letter of the AI ecosystem pillar followed by a number, i.e., the first guideline under Human will be [H1].

## Humans

In this section, we cover guidelines for all the elements that are related to: (1) enablers for humans to engage with AI, and (2) all the impacts humans have on the AI ecosystem. We will refer to stakeholders as including, but not limited to, AI users, AI project team members, employers, commissioning organizations, government regulatory bodies, legislators, civil society organizations monitoring AI impact and advocating for users' rights, community organizations, industry, and people affected by AI systems.

**[H1]  Incorporate diversity and inclusion principles and practice throughout the AI lifecycle**

Integrating diversity and inclusion principles and practices by engaging diverse stakeholders throughout the lifecycle of AI has an important role in achieving equity for all stakeholders [25]. End-users, AI practitioners, subject matter experts, investors, and other professionals including those from the law, social sciences, and community development, should be involved to identify potential impacts.

**[H2]  Identify stakeholder knowledge and needs**

Stakeholders generally hold specific knowledge, expertise, concerns, and objectives that can contribute to effective AI system design, development, and deployment. These include government regulatory bodies, and civil society organizations monitoring AI impact, industry, and people affected by AI systems. There are groups whose diversity of knowledge or expertise is valuable for AI system design, but they do not necessarily have needs or requirements for the system because they will not be users or consumers, however, they should be consulted.

### [H3] Examine key questions collaboratively, such as AI's purpose, intended users, and creators

Key questions about why an AI project should happen, for who is the project for, and by whom it should be developed should be asked, answered, and revisited collectively using a diversity and inclusion lens during the AI-LC. Views from stakeholders and representatives of impacted communities should be sought. Although it might be advantageous that some AI design team members are themselves representative of impacted stakeholders and thus well-placed to recognize and address potential inclusion-related harms, diverse team members should not be expected to stand in for impacted cohorts (whom themselves would hold a diversity of perspectives, expectations, and experiences).

### [H4] Implement inclusive and transparent feedback mechanisms for stakeholders

Users should have accessible mechanisms to identify and report harmful or concerning AI system incidents and impacts, with such warnings shareable among relevant stakeholders [26] [27]. Provide equally accessible, safe, and anonymous mechanisms for internal and external stakeholders (including employees of a commissioning organization or AI development company) to express their concerns without fear. Feedback should be continuously incorporated into system updates and communicated to relevant stakeholders.

### [H5] Monitor and adapt to changes in the operating context

Processes to identify and respond to changes in the operating context, including the potential appearance of new user groups who may be treated differentially by the AI system, should be established. For example, a computational medical system trained in large metropolitan hospitals may not work as intended when used in small rural hospitals due to various factors including training of local healthcare personnel, quality of clinical data entered the system, or behavioral factors affecting how humans interact with AI [2].

### [H6] Adopt a socio-technical approach to create human-centered AI

An approach to human-in-the-loop that considers a broad set of socio-technical factors should be adopted. Relevant fields of expertise include human factors, psychology, organizational behavior, and human-AI interaction. However, researchers from Stanford University argue that "practitioners should focus on AI in the loop", with humans remaining in control. They advise that "all AI systems should be designed for augmenting and assisting humans – and with human impacts at the forefront." So they advocate the idea of "human in charge" rather than human in the loop [10].

### [H7]  Create AI literacy and education programs

An 'AI-ready' person is someone who knows enough to decide how, when and if they want to engage with AI. Critical AI literacy is the pathway to such agency. Consequently, governments should drive the equitable development of AI-related skills to everyone from the earliest years via formal, informal, and extracurricular education programs covering technical and soft skills, along with awareness of digital safety and privacy issues. Governments and civil society organizations should create, and fund grant schemes aimed at enhancing the enrolment of women in AI education. Organizations also can play a critical role via paid internships and promoting community visits, talks, workshops, and engagement with AI practitioners. To harness the potential of increasing diversity and inclusion in the global AI ecosystem, such opportunities should prioritize participation (as facilitators and participants) of people with diverse attributes (including cultural, ethnic, age, gender identification, cognitive, professional, etcetera).

### [H8]  Prioritize equitable hiring practices & career development opportunities

Data science teams should be as diverse as the populations that the built AI systems will affect. Product teams leading and working on AI projects should be diverse and representative of impacted user cohorts. Diversity, equity, and inclusion in the composition of teams training, testing and deploying AI systems should be prioritized as the diversity of experience, expertise, and backgrounds is both a critical risk mitigant and a method of broadening AI system designers' and engineers' perspectives. For example, female-identifying role models should be fostered in AI projects [11]. Diversity and inclusion employment targets and strategies should be regularly monitored and adjusted if necessary.

The WEF Blueprint recommends four levers [2]. First, widening career paths by employing people from non-traditional AI backgrounds, embedding this goal in strategic workplace planning. For instance, backgrounds in marketing, social media marketing, social work, education, public health, and journalism can contribute fresh perspectives and expertise. Second, diversity and inclusion should be covered in training and development programs via mentorships, job shadowing, simulation exercises, and contact with diverse end user panels. Third, partnerships with academic, civil society and public sector institutions should be established to contribute to holistic and pan-disciplinary reviews of AI systems, diversity and inclusion audits, and assessment of social impacts. Fourth, a workplace culture of belonging should be created and periodically assessed via both open and confidential feedback mechanisms which include diversity markers.

### [H9]  Operationalize inclusive and substantive community engagement

A vast body of knowledge about community engagement praxis exists. Guidelines and frameworks are updated and operationalized by practitioners from many disciplines including community cultural development, community arts, social work, social sciences, architecture, and public health. However, this vital element is largely neglected in the AI ecosystem although many AI projects would benefit from considered attention to community engagement. For instance, in the health sector, AI and advanced analytics implementation in primary care should be a collaborative effort that involves patients and communities from diverse social, cultural, and economic backgrounds in an intentional and meaningful manner [12].

A Community Engagement Manager role could be introduced who would work with impacted communities throughout the AI-LC and for a fixed period post-deployment. Reciprocal and respectful relationships with impacted communities should be nurtured, and community expectations about both the engagement and the AI system should be defined and attended to. If impacted communities contain diverse language, ethnic, and cultural cohorts a Community Engagement Team from minority groups would be more appropriate. One role would be to develop tailored critical AI literacy programs for example. Organizations must put "the voices and experiences of those most marginalized at the center" when implementing community engagement outcomes in an AI project [2].

## Data

Data is at the heart of Artificial Intelligence. What, how, why, by whom and for whom data is collected, labelled, modelled, stored, and applied has many implications for diversity and inclusion, and increasingly so in the era of Big Data and Machine Learning. In this section, we suggest some general guidelines that can be adapted to suit the specific context of an AI project.

### [D1]     Establish a clear rationale for data collection

For data collection involving human subjects, why, how and by whom data is being collected should be established in the Pre-Design stage. Data literacy is needed for all those dealing with data. The validity and accuracy of the data must be ensured. Key stakeholders and data scientists should identify potential data challenges or data biases that have diversity and inclusion implications. Project teams should develop mitigation and monitoring strategies and systematically capture and regularly review data bias issues.

**[D2] Empower stakeholders and other knowledge holders in data selection, collection, and analysis, to ensure demographic representation**

Representatives of impacted stakeholders should be identified and partnered with on data collection methods. This is particularly important when identifying new or non-traditional data-gathering resources and methods. To increase representativeness and responsible interpretation, when collecting and analyzing specific datasets include diverse viewpoints and different kinds of knowledge holders. Attention to agency and autonomy, informed consent, and transparency, to ensure a safe environment for data collection from representatives of vulnerable groups, is essential. The sensitive nature of some attributes (such as sexuality or religion) means that some people might not feel comfortable disclosing their attributes. Training data sets should be demographically representative of the cohorts or communities that will be impacted by the AI system.

**[D3] Establish procedures for data privacy and offer opt-out options**
Data privacy should be at the forefront, particularly when data from marginalized populations are involved. End users should be offered choices about privacy and ethics in the collection, storage, and use of data. Opt-out methods for data collected for model training and model application should be offered where possible.

**[D4] Adhere to data sovereignty principles and practices**
The concept of, and practices supporting, data sovereignty is a critical element in the AI ecosystem. It covers considerations of the "use, management and ownership of AI to house, analyze and disseminate valuable or sensitive data" [2]. Although definitions are context-dependent, operationally data sovereignty refers to stakeholders within an AI ecosystem, and other relevant representatives from outside stakeholder cohorts to be included as partners throughout the AI-LC. Data sovereignty should be explored from and with the perspectives of those whose data is being used. These alternative and diverse perspectives can be captured and fed back into AI Literacy programs, exemplifying how people can affect and enrich AI both conceptually and materially.

Various Indigenous technologists, researchers, artists, and activists have progressed the concept of, and protocols for, Indigenous data sovereignty in AI. This involves "Indigenous control over the protection and use of data that is collected from our communities, including statistics, cultural knowledge and even user data," and moving beyond the representation of impacted users to "maximising the generative capacity of truly diverse groups" [13].

**[D5]    Acknowledge the connections between access issues, infrastructure, capacity building, and data sovereignty**
Access, including cloud and offline data hosting, should be attended to because government and industry generally build and manage these on their own terms. Access is directly connected to capacity building (teams and stakeholders) and data sovereignty issues.

**[D6]    Assess dataset suitability factors**
Dataset suitability factors should be assessed, including cultural diversity considerations. This also includes statistical methods for mitigating representation issues, the socio-technical context of deployment, and the interaction of human factors with the AI system. The question of whether suitable datasets exist that fit the purpose of the various applications, domains, and tasks for the planned AI system should be asked. Other relevant questions to be considered are: who decides on dataset suitability, how is data defined, and how to deal with unavailable data.

**[D7]    Document social descriptors when scraping data from different sources and perform compatibility analysis**
Developers should attend to and document the social descriptors (for example, age, gender, and geolocation) when scraping data from different sources including websites, databases, social media platforms, enterprise applications, or legacy systems. Context is important when the same data is later used for different purposes such as asking a new question about an existing data set [14]. A compatibility analysis should be performed to ensure that potential sources of bias are identified, and mitigation plans made. This analysis would capture context shifts in new uses of data sets, identifying whether or how these could produce specific bias issues.

**[D8]    Enhance feature-based labeling and develop precise user identity notions**
Apply more inclusive and socially just data labelling methodologies such as Intersectional Labeling Methodology to address gender bias [23]. Rather than relying on static, binary gender in a face classification infrastructure, application designers should embrace and demand improvements, to feature-based labelling. For instance, labels based on neutral performative markers (e.g. beard, makeup, dress) could replace gender classification in the facial analysis model, allowing third parties and individuals who come into contact with facial analysis applications to embrace their own interpretations of those features. Instead of focusing on improving methods of gender classification, application designers could use labelling alongside other qualitative data such as Instagram captions to formulate more precise notions about user identity [15].

**[D9] Consider context drift and data transparency**

Observed context drift in data should be documented via data transparency mechanisms capturing where and how the data is used and its appropriateness for that context. Harvard researchers have expanded the definition of data transparency, noting that some raw data sets are too sensitive to be released publicly, and incorporating guidance on development processes to reduce the risk of harmful and discriminatory impacts:
- "In addition to releasing training and validation data sets whenever possible, agencies shall make publicly available summaries of relevant statistical properties of the data sets that can aid in interpreting the decisions made using the data, while applying state-of-the-art methods to preserve the privacy of individuals.
- When appropriate, privacy-preserving synthetic data sets can be released in lieu of real data sets to expose certain features of the data if real data sets are sensitive and cannot be released to the public." [5]

Teams should use transparency frameworks and independent standards; conduct and publish the results of independent audits; open non-sensitive data and source code to outside inspection [7].

## Process

In this section, we suggest guidelines both for the main stages of the AI system development lifecycle, and those that specifically respond to the AI ecosystem's evolutionary and iterative qualities. The guidelines draw on long-standing principles and practices in the Human-Centered Design field, and in some cases adapted for elements pertaining to AI.

### *Pre-Development Process*

**[P1]    Practise inclusive problem identification and impact assessment**
A project owner (individual or organization) with suitable expertise and resources to manage an AI system project should be identified, ensuring that accountability mechanisms to counter potential harm are built in. It should be decided which other stakeholders will be involved in the systems development and regulation. Both intended and unintended impacts that the AI system will or might have should be assessed in collaboration with stakeholder communities, with additional experts being consulted if necessary.

Incorporate inclusive tech principles such as normalizing inclusion at a systemic level; designing with excluded and diverse communities, not for them; promoting accountability; enforcing data governance to ensure ethical practices are being met [7].

**[P2]    Establish mechanisms for monitoring and improvement**
Mechanisms enabling an iterative process of continuous monitoring and improvement of diversity and inclusion considerations should be established from the outset. These will help ensure that all stakeholders' needs are met and that inadvertent harm is not caused. Both team and system performance should be regularly assessed, improvements identified, and changes executed accordingly.

**[P3]    Identify possible systemic problems of bias and designate a responsible board**
At the start of the Design stage, stakeholders should identify possible systemic problems of bias such as racism, sexism, or ageism that have implications for diversity and inclusion. Main decision-makers and power holders should be identified, as this can reflect systemic biases and limited viewpoints within the organization.

A board responsible for algorithmic bias should be appointed. This role entails broad oversight over strategic decisions and accountability for mitigating bias (in consultation with team and stakeholders) [16].

**[P4]    Consider specific labelling categories relevant to the AI system**
For example, before embedding gender classification into a facial analysis service or incorporating gender into image labelling, it is important to consider what purpose gender is serving. Furthermore, it is important to consider how gender will be defined, and whether that perspective is unnecessarily exclusionary (for example, non-binary). Therefore, stakeholders involved in the development of facial analysis services and image datasets should assess the potentially negative and harmful consequences their service might be used for—including emotional, social, physical, and systematic (state or governmental) harms [15].

*Development Process*

**[P5]    Evaluate multiple trade-offs in AI development**
In the design stage, decisions should weigh the social-technical implications of the multiple trade-offs inherent in AI systems. These trade-offs include the system's predictive accuracy which is measured by several metrics. The metrics include accuracies within sub-populations or across different use cases, as partial and total accuracies. Fairness outcomes for different sub-groups of people the AI systems will be applied to or made decisions for. The other trade-offs could be related to generalizability, interpretability, transparency or explainability.

Acknowledge the challenges of trading off and balancing fairness and accuracy especially when they influence high-stake decisions. For instance, in the field of computational medicine, post-hoc correction methods based on randomizing predictions that are unjustifiable from an ethical perspective in clinical tasks (for example, severity scoring) should be avoided [24].

Teams should decide how to treat "multiple axes of identities" in the machine learning pipeline to reduce the risk of unfairness or harm. Attention to intersectionality throughout the AI-LC ranges from selecting which identity labels to use in datasets, decisions about how to "technically handle the progressively smaller number of individuals in each group that will result from adding additional identities and axes" during model training and deciding how to perform fairness evaluation as the number of groups increases [17].

### [P6]    Create model designs with diversity and inclusion in mind

Diverse values and cultural perspectives from multiple stakeholders and populations should be codified in mathematical models and AI system design. Basic steps should include incorporating input from diverse stakeholder cohorts, ensuring the development team embodies different kinds of diversity, establishing and reviewing metrics to capture diversity and inclusion elements throughout the AI-LC, and ensuring well-documented end-to-end transparency on final design choices.

### [P7]    Evaluate, modify, and document bias identification and mitigation measures

During model training and implementation the effectiveness of bias mitigation should be evaluated and adjusted. Periodically assess bias identification processes and address any gaps. The model specification should include how and what sources of bias were identified, mitigation techniques used, and how successful mitigation was. A related performance assessment should be undertaken before model deployment.

### [P8]    Develop effective validation processes

Subject matter experts should create and oversee realistic validation processes addressing bias-related challenges including noisy labelling (for example, mislabeled samples in training data), use of proxy variables, and performing system tests under optimal conditions unrepresentative of real-world deployment context.

### [P9]    Apply Value Sensitive Design principles and methodology

Teams should engage with the complexity in which people experience values and technology in daily life. Values should be understood holistically and as being interrelated, rather than being analyzed in isolation from one another.

### [P10] Design evaluation tasks that best mirror the real-world setting

Evaluation, even on crowdsourcing platforms used by ordinary people, should capture end users' types of interactions and decisions. The evaluations should demonstrate what happens when the algorithm is integrated into a human decision-making process. Does that alter or improve the decision and the resultant decision-making process as revealed by the downstream outcome?

### [P11] Apply fairness analysis throughout the development process

Rather than thinking of fairness as a separate initiative, it's important to apply fairness analysis throughout the entire process, making sure to continuously re-evaluate the models from the perspective of fairness and inclusion [18]. The use of Model Performance Management tools or other methods should be considered to identify and mitigate any instances of intersectional unfairness [19]. For example, a diversity rating audit that combines various attributes including age, gender, and ethnicity can be used to audit data sets used to train AI algorithms [8].

### [P12] Assess the applicability of Human-centered design (HCD) methodology for AI system development

A Human-centered design (HCD) methodology, based on International Organization for Standardization (ISO) standard 9241-210:2019, for the development of AI systems, could comprise:

- Defining the Context of Use, including operational environment, user characteristics, tasks, and social environment;
- Determining the User & Organizational Requirements, including business requirements, user requirements, and technical requirements;
- Developing the Design Solution, including the system design, user interface, and training materials; and
- Conducting the Evaluation, including usability and conformance testing [1].

### [P13] Establish diverse partnerships and training populations

Engage with domain experts, ethicists, and antiracism experts among others in developing, training, testing, and implementing models. Recruit diverse and representative populations in training samples [20].

## *Post-Development Process*

### [P14]   Conduct monitoring and evaluation during deployment

New or emergent stakeholder cohorts should participate in monitoring the system for bias and retraining when necessary. Stakeholders should be involved in a final review and sign-off, particularly if their input propelled significant changes in design or development processes. After validation, teams should obtain informed consent on the developed product features from impacted stakeholders, to track and respond to the system's impact on different communities.

When the AI system will have a direct impact on citizens then public communication regarding possible impacts on lives or services should occur. Announcements should be presented in multiple languages via a range of media to reach the widest possible audience. It is important to have mechanisms for managing expectations after deployment.

### [P15]   Undertake holistic external impact monitoring

AI systems' learning capabilities evolve. External contexts such as climate, energy, health, economy, environment, political circumstances, and operating contexts also change. Therefore, both AI systems and the environment in which they operate should be continuously monitored and reassessed using appropriate metrics and mitigation processes, including methods to identify the potential appearance of new user groups who may be treated differentially by the AI system. Teams should consider the entire decision-making process not only the algorithm in isolation. Even if an algorithm satisfies the criteria of fair or accurate and is deemed not risky, it still can have downstream consequences when deployed with human interactions. Software tools monitoring system behavior should be complemented by teams who can assess and respond to impacted stakeholders.

Detailed policies and procedures on how to handle system output and behavior should be developed and followed. Observed deviations from goals should trigger feedback loops and subsequent adjustments to data curation and problem formulation in the model, followed by further continuous testing and evaluation.

### [P16]   Monitor and audit changing AI system impacts

It is critical to monitor the use of advanced analytics and AI technology to ensure that benefits are accruing to diverse groups in an equitable manner [12]. The scale of AI system impact can change rapidly and unevenly when deployed.

Organizations should build resilience, flexibility, and sensitivity to respond to changes to ensure equitable and inclusive outcomes.

### [P17] Test and evaluate bias characteristics during deployment
The deploying organization and other stakeholders should use documented model specifications to test and evaluate bias characteristics during deployment in the specific context.

### [P18] Collect demographic data from users to assist in bias monitoring
Monitoring for bias should collect demographic data from users including age and gender identity to enable the calculation of assessment measures.

## System

AI systems have been defined and classified in different ways. For example, one classification is based on the methods used for development, such as symbolic AI (using Logic), probabilistic inference (using Bayesian networks), and connectionist (based on the human brain). Current AI technology is typically a system that for a given set of human-defined objectives, typically uses large historical datasets to learn, and make predictions, recommendations, or decisions for humans or for other larger systems where AI is a component. In this section, we provide a few guidelines for the AI system in its context of use.

### [S1] Establish inclusive and informed product development, training, evaluation, and sign-off
New stakeholders for iterative rounds of product development, training, and testing should be brought in, and beta groups for test deployments should be recruited. User groups should reflect different needs and abilities. Fresh perspectives contribute to the evaluation of both the AI system's functionality and, importantly, its level and quality of inclusivity. New or emergent stakeholder cohorts should participate in system monitoring and retraining. Stakeholders should be involved in a final review and sign-off, particularly if their input propelled significant changes in design or development processes. After validation, teams should obtain informed consent on the developed product features from impacted stakeholders, to track and respond to the system's impact on different communities.

### [S2] View AI systems through a holistic lens
Code is not the right level of abstraction at which to understand AI systems, whether it is for accountability or adaptability. Instead, systems should be analyzed in terms of inputs and outputs, overall design, embedded values, and how the software system fits with the overall institution deploying it [21].

### [S3]   Employ model design techniques to address diversity and inclusion considerations

Diverse values and cultural perspectives from multiple stakeholders and populations should be codified in mathematical models and AI system design. Model design techniques are necessarily contextual, related to the type of AI technology, the purpose and scope of the system, how users will be impacted, and so forth. However, basic steps should include incorporating input from diverse stakeholder cohorts, ensuring the development team embodies different kinds of diversity to reflect the diversity of AI system end-users and stakeholders, establishing and reviewing metrics to capture diversity and inclusion elements throughout the AI-LC, and ensuring well-documented end-to-end transparency on final design choices.

### [S4]   Evaluate, adjust, and document bias identification and mitigation measures

During model training and implementation the effectiveness of bias mitigation should be evaluated and adjusted. Periodically assess bias identification processes and address any gaps. The model specification should include how and what sources of bias were identified, mitigation techniques used, and how successful mitigation was. A related performance assessment should be undertaken before model deployment.

### [S5]   Consider context issues during model selection and development

Context should be taken into consideration during model selection to avoid or limit biased results for sub-populations. Caution should be taken in systems designed to use aggregated data about groups to predict individual behavior as biased outcomes can occur. "Unintentional weightings of certain factors can cause algorithmic results that exacerbate and reinforce societal inequities," for example, predicting educational performance based on an individual's racial or ethnic identity [1].

## Governance

Governance is defined as "a framework of policies, rules, and processes for ensuring direction, management and accountability" [1]. In the larger AI ecosystem governance occurs at organizational, industry, and project team levels. In this section we include only those governance considerations most relevant to AI project teams because this is the area over which data scientists, team leaders, managers, and stakeholders have direct control.

### [G1] Establish policies for biometric data collection and usage

Establishing policies (either at the organizational or industry level) that comply with Privacy Acts and other data collection and data use legislation, for how biometric data and face and body images are collected and used may be the most effective way of mitigating harm to trans people, marginalized races, ethnicities, and sexualities [15].

### [G2] Triage and tier AI bias risks

AI is not quarantined from negative societal realities such as discrimination and unfair practices. Consequently, it is arguably impossible to achieve zero risk of bias in an AI system. Therefore, AI bias risk management should aim to mitigate rather than avoid risks. Risks can be triaged and tiered; resources allocated to the most material risks, the worst problems and most sensitive uses, those "most likely to cause real-world harm" [1] [21].

### [G3] Align AI bias mitigation efforts with relevant legislation

Bias mitigation should be aligned with relevant existing and emerging legal standards. This includes national and state laws covering AI use in hiring, eligibility decisions (e.g., credit, housing, education), discrimination prohibitions (e.g., race, gender, religion, age, disability status), privacy, and unfair or deceptive practices.

### [G4] Adhere to AI risk assessment frameworks

Teams should develop diversity and inclusion policies and procedures addressing key roles, responsibilities, and processes within the organizations that are adopting AI. Bias risk management policies should specify how risks of bias will be mapped and measured, and according to what standards.

AI risk practice and associated checks and balances should be embedded and ingrained throughout all levels of the relevant stakeholder organizations. This may require a cultural shift while the AI system is evolving, and an acceptance that neither all questions will be answered necessarily, nor all problems well understood [21].

AI risk mitigation should not be framed as a quantitative balance but rather as an interdisciplinary qualitative judgement. Stakeholders should consider vetoing a deployment if they foresee potential unintended consequences [21].

The iterative and continuous AI risk assessment process should be adopted and understood as contributing to an organizational cultural shift. Risk management should be viewed as being beneficial to the organization. The use of methods not necessarily common in the computer science field such as story telling could be explored [21].

Regular risk assessment for diversity and inclusion in an AI system should assess the following points:

- which practices have emerged to date
- which practices seem no longer relevant
- if the initial vision for diversity and inclusivity has been achieved or on-track
- if the use case and user group have been equitably defined or require refining
- whether a proactive approach to inclusivity has been adequately prioritized throughout the development process
- whose perspectives are or were over-represented or under-represented
- if any unforeseen challenges or activities arose and if so, what measures were taken in response
- how the development process has further informed the understanding of any protected or at-risk groups [2].

## [G5] Implement inclusive tech governance practices

Organizations should Implement responsible AI leadership, drawing on existing resources such as UC Berkeley's Equity Fluent Leadership Playbook [22]. They should engage personnel to implement and monitor compliance with AI ethics principles, and train leaders to operationalize AI and data governance and measure engagement [7]. The governance mechanisms/guidelines should be connected with lower-level development/design patterns. E.g., risk assessment framework can be supported by continuous risk assessment component in the AI ecosystem.

## [G6] Establish inclusive AI Infrastructure

An inclusive AI ecosystem involving the broadest range of community members requires equitable access to technical infrastructure (computing, storage, networking) to facilitate the skilling of new AI practitioners and offer opportunities for citizens' development of AI systems. Both governments and industry should create infrastructure, data management and knowledge-sharing protocols, and education programs. They should work with civil society organizations to support national and global networks.

# Conclusion

Diversity and Inclusion in AI systems promote a humanist view of product development. Humans must be placed at the center of the AI ecosystem, whether it is within the process (entire AI development lifecycle), for the product (in the context of system use and for those impacted directly or indirectly by AI technology), in data (fair representation of all relevant stakeholders in the datasets), or in governance (legal frameworks, regulations, policies, and guidelines).

The DI-AI guidelines presented in this chapter advise technologists and other professionals in the AI field on *what* should be done to ensure that diversity and inclusion factors are adequately considered in decision-making, software development, and risk assessment throughout the AI lifecycle. At times we also suggest *who* should be responsible for oversight, implementation, monitoring, and evaluation of the work entailed in the specific guidelines. Various methods exist that would enable the guidelines to be implemented. For instance, the Pattern Catalogues presented in this book can be adapted to many of the DI-AI guidelines to suggest *how* stakeholders and teams can implement them.

The DI-AI guidelines in this chapter have been organized and presented around the 5 pillars of the DI-AI definition, while the Responsible AI Pattern Catalogue in the rest of this book has been organized differently and references Governance, Process, Product, and Techniques. Moreover, both the DI-AI guidelines and patterns are deliberately open and non-prescriptive to allow them to be used, re-used, and adapted to suit the *context and scale* of any AI system. Consequently, those in charge of Responsible AI within an organization, project, and/or team can consider a many-to-many mapping of the guidelines and patterns that fits their own purposes. We offer a few examples below under the 5 DI-AI pillars.

**Human**
**Incorporate diversity and inclusion principles and practice throughout the AI lifecycle [H.1] -** This guideline recommends that the composition of different levels of stakeholder cohorts should maintain diversity along social lines and that end-users, AI practitioners, subject matter experts, and professionals should be involved to identify downstream impacts comprehensively. It can be mapped to **Stakeholder Engagement [G.20]** in the organization-level governance patterns, and to the **Code of RAI [G.11]** at industry-level governance. Resultant benefits include increased stakeholder trust in the AI project, reduced risk, clear guidance for employees, and the same explicit rules for everyone in the organization.

**Data**
**Empower stakeholders and other knowledge holders in data selection, collection, and analysis, to ensure demographic representation [D.2] -** Diverse viewpoints, including those of 'non-experts', should be elicited when collecting and analyzing specific datasets to identify potential problems and risks. This guideline can be mapped to **Stakeholder Engagement [G.20]**, and to the technique/product patterns **Fairness Assessor [T.1]** and **Discrimination Mitigator [T.2].** Fairness metrics could include demographic parity and equal opportunity, and algorithmic discrimination can be addressed by various pre-processing, in-processing and post-processing techniques.

**Process**
**Assess the applicability of Human-centered design (HCD) methodology for AI system development [P.12] -** This guideline could be addressed by following 3 requirements stage process patterns: **Verifiable Ethical Requirement [P.2]**, **Data Requirement [P.2]**, and **Ethical User Story [P.4]**. Business analysts would drive the first two patterns. They would be joined by product managers, AI users, and AI consumers in the creation of ethical user stories.

**System**
**Employ model design techniques attuned to diversity and inclusion considerations [S.3]** – Diverse values and cultural perspectives from multiple stakeholders and populations should be codified in mathematical models and AI system design. The design stage process pattern **RAI Design Modelling [P.7]** details relevant practical steps for AI architects including designing formal models aligned with human values, creating RAI knowledge bases to inform design decisions that consider ethical concerns, and using logic programming to implement ethical principles.

**Governance**
**Triage and tier AI bias risks [G.2] -** AI bias risk management should aim to mitigate rather than avoid risks. Resources should be allocated to the most material risks, the worst problems, and the most sensitive uses. The organization-level governance pattern **RAI Risk Assessment [G.12]** is relevant to this DI-AI guideline by aiming at management teams and covers both domain-specific risks and emerging risks in constantly evolving AI systems. It recommends that an RAI risk assessment framework be co-designed with key stakeholders including a RAI risk committee, development teams, and prospective purchasers.

In summary, Diversity and Inclusion must occupy a special place in the AI ecosystem. Its practices must be acknowledged and valued in all aspects of the AI system development life cycle, rather than be reduced to merely as one of the practices implicitly inferred from the overall understanding of achieving "fairness" in AI. We advocate that Diversity and Inclusion principles should be at the core of AI ethical principles and embedded in the design, development, deployment, and evolution of all AI systems. We believe that this embedding will in turn accelerate and improve our understanding and practices of diversity and inclusion in society.